\documentclass{article}
\usepackage{microtype}
\usepackage{graphicx}
\usepackage{booktabs}
\usepackage{amsmath,amssymb}
\usepackage[accepted]{mlsys2025}
\usepackage{algorithm}
\usepackage{algorithmic}
\usepackage[hyphens]{url}
\usepackage{hyperref}
\usepackage{xspace}
\definecolor{paperlink}{rgb}{0,0.08,0.45}
\hypersetup{colorlinks=true,linkcolor=paperlink,citecolor=paperlink,urlcolor=paperlink,pdfauthor={Wentao Ni, Hong Cai, Mingu Lee, Chihao Yu, Zhongming Yu, Hejia Zhang, Kangqi Zhang, Fatih Porikli, Zhijian Liu, Jishen Zhao},pdftitle={TileWeave: Heterogeneous Tiling for Sparse Prefill on Mobile SoCs},pdfkeywords={On-device inference, heterogeneous computing, sparse attention, KV cache, Snapdragon},pdfsubject={TileWeave preprint}}
\newcommand{\prefillrows}{%
512 & 1029 / 1031 & 775 / 780 & 766 / 765 & 2441 / 2310 & 2384 / 2350 & 2525 / 2472 \\
1024 & 2713 / 2716 & 1716 / 1706 & 1417 / 1404 & 2455 / 2208 & 2488 / 2350 & 2715 / 2428 \\
2048 & 3826 / 3820 & 2011 / 2017 & 1592 / 1589 & 2311 / 2001 & 2445 / 2188 & 2657 / 2256 \\
4096 & 6185 / 6210 & 2658 / 2664 & 1999 / 1993 & 1996 / 1730 & 2310 / 2028 & 2327 / 2029 \\
8192 & 11982 / 12028 & 3917 / 3919 & 3123 / 3120 & 1474 / 1340 & 1965 / 1779 & 1911 / 1771 \\%
}

\newcommand{\existingselectorrows}{%
Dense & -- & -- & -- & 1.00 / 1.00 & -- & 87.10 & -- \\
\midrule
XAttention & antidiagonals & 12 / 14 & 0.7 / 1.0 & 1.14 / 1.05 & 1.57 / 1.58 & 86.39 & 86.46 (62.2) \\
MInference (vertical-slash) & last 64 queries & 20 / 28 & 0.5 / 0.6 & 1.35 / 1.46 & 2.20 / 2.67 & 83.47 & 87.07 (98.6) \\
FlexPrefill & last 128 queries & 43 / 58 & 1.1 / 1.5 & 1.37 / 1.40 & 1.43 / 1.43 & 81.28 & 85.59 (62.9) \\
SampleAttention & last 128 queries & 38 / 54 & 0.4 / 0.5 & 1.32 / 1.35 & 1.39 / 1.39 & 81.22 & 84.32 (62.9) \\
SpargeAttn & self-similarity & 23 / 28 & 1.1 / 1.2 & 1.16 / 1.09 & 1.20 / 1.13 & 63.28 & 66.25 (61.9) \\
MInference (block-sparse) & pooled Q, K & 5 / 5 & 0.5 / 0.4 & 1.02 / 0.82 & 1.41 / 1.11 & 75.04 & 77.82 (58.9) \\
FlashPrefill & pooled K & 8 / 9 & 0.1 / 0.1 & 1.00 / 0.81 & 1.06 / 0.86 & 83.10 & 84.38 (58.8) \\%
}

\newcommand{\offloadallmsfour}{548.9}
\newcommand{\offloadonemsfour}{453.8}
\newcommand{\offloadtwomsfour}{367.4}
\newcommand{\offloadthreemsfour}{289.5}
\newcommand{\offloadonespeedfour}{1.21}
\newcommand{\offloadtwospeedfour}{1.49}
\newcommand{\offloadthreespeedfour}{1.90}
\newcommand{\offloadallmseight}{1981.4}
\newcommand{\offloadonemseight}{1590.0}
\newcommand{\offloadtwomseight}{1246.0}
\newcommand{\offloadthreemseight}{934.7}
\newcommand{\offloadonespeedeight}{1.25}
\newcommand{\offloadtwospeedeight}{1.59}
\newcommand{\offloadthreespeedeight}{2.12}
\newcommand{\motivationqualityrows}{%
Pool + diagonal & 85.57 & 55.42 \\
Pool + exceptions (ours) & \textbf{86.32} & 62.19 \\
Union & 86.29 & 73.13 \\%
}
\newcommand{\motivationqualitygain}{0.75}
\newcommand{\motivationqualitycilow}{0.37}
\newcommand{\motivationqualitycihigh}{1.14}

\newcommand{\selectionrows}{%
XAttention antidiagonal & 7505 & 1.0$\times$ \\
Full Q mean & 2489 & 3.0$\times$ \\
Sampled Q mean (8 rows) & 727 & 10.3$\times$ \\
Sampled Q mean (4 rows) & 378 & 19.9$\times$ \\%
}

\newcommand{\rulershortrows}{%
Dense & 97.30 & 100 & 91.13 & 100 \\
XAttention & 96.88 & 86.3 & 90.73 & 68.9 \\
TileWeave & 97.00 & 82.1 & 90.69 & 69.2 \\%
}

\newcommand{\rulerablationrows}{%
\multicolumn{6}{l}{\emph{Attention (ms)}} \\
XAttention & naive, per K/V head & 3,762 & 2.24$\times$ & 12,978 & 2.37$\times$ \\
XAttention & naive, per head & 7,263 & 4.33$\times$ & 25,207 & 4.61$\times$ \\
Pooled keys & naive & 2,552 & 1.52$\times$ & 8,892 & 1.63$\times$ \\
Pooled keys & split, $c^*{=}2$ & 1,826 & 1.09$\times$ & 6,007 & 1.10$\times$ \\
Pooled keys & split, $c^*{=}3$ & 1,677 & -- & 5,470 & -- \\
\midrule
\multicolumn{6}{l}{\emph{Prefill (s)}} \\
XAttention & naive, per K/V head & 6.61 & 1.48$\times$ & 18.78 & 1.69$\times$ \\
XAttention & naive, per head & 10.12 & 2.27$\times$ & 30.95 & 2.79$\times$ \\
Pooled keys & naive & 5.50 & 1.24$\times$ & 14.84 & 1.34$\times$ \\
Pooled keys & split, $c^*{=}2$ & 4.61 & 1.03$\times$ & 11.66 & 1.05$\times$ \\
Pooled keys & split, $c^*{=}3$ & 4.45 & -- & 11.10 & -- \\%
}

\newcommand{\executionrows}{%
TileWeave & 82.98 & 58.0 & 2,375 & 7,864 & 6.77 & 16.60 \\
Masked tile & 82.64 & 51.1 & 3,066 & 10,537 & 7.63 & 19.63 \\
Union & 83.16 & 70.1 & 2,949 & 10,092 & 7.52 & 19.18 \\
\midrule
\multicolumn{3}{l}{Speedup over the masked tile} & 1.29$\times$ & 1.34$\times$ & 1.13$\times$ & 1.18$\times$ \\
\multicolumn{3}{l}{Speedup over the union} & 1.24$\times$ & 1.28$\times$ & 1.11$\times$ & 1.16$\times$ \\%
}

\newcommand{\prefilllatencyrows}{%
\multicolumn{6}{l}{\emph{SM8750}} \\
Dense & 0.48 & 1.00 & 2.37 & 6.09 & 17.53 \\
XAttention & 0.63 & 1.51 & 3.81 & 10.12 & 30.95 \\
ShadowAttn & 1.04 & 3.49 & 15.36 & 92.90 & 539.11 \\
ShadowAttn (GPU) & 1.09 & 3.26 & 10.64 & 36.30 & 139.74 \\
TileWeave & 0.47 & 0.95 & 2.02 & 4.45 & 11.10 \\
\midrule
vs.\ Dense & 1.01$\times$ & 1.05$\times$ & 1.17$\times$ & 1.37$\times$ & 1.58$\times$ \\
vs.\ XAttention & 1.35$\times$ & 1.59$\times$ & 1.88$\times$ & 2.27$\times$ & 2.79$\times$ \\
vs.\ ShadowAttn & 2.21$\times$ & 3.68$\times$ & 7.60$\times$ & 20.86$\times$ & 48.56$\times$ \\
vs.\ ShadowAttn (GPU) & 2.31$\times$ & 3.44$\times$ & 5.26$\times$ & 8.15$\times$ & 12.59$\times$ \\
\midrule
\multicolumn{6}{l}{\emph{SM8850}} \\
Dense & 0.39 & 0.79 & 1.79 & 4.22 & 11.58 \\
XAttention & 0.53 & 1.24 & 3.16 & 8.43 & 25.96 \\
ShadowAttn & 1.19 & 3.52 & 13.02 & 77.05 & 508.27 \\
ShadowAttn (GPU) & 0.76 & 2.17 & 7.06 & 24.59 & 112.39 \\
TileWeave & 0.37 & 0.73 & 1.53 & 3.29 & 8.04 \\
\midrule
vs.\ Dense & 1.03$\times$ & 1.08$\times$ & 1.17$\times$ & 1.29$\times$ & 1.44$\times$ \\
vs.\ XAttention & 1.41$\times$ & 1.70$\times$ & 2.06$\times$ & 2.56$\times$ & 3.23$\times$ \\
vs.\ ShadowAttn & 3.20$\times$ & 4.82$\times$ & 8.49$\times$ & 23.44$\times$ & 63.23$\times$ \\
vs.\ ShadowAttn (GPU) & 2.02$\times$ & 2.97$\times$ & 4.61$\times$ & 7.48$\times$ & 13.98$\times$ \\%
}

\newcommand{\attentionlatencyrows}{%
\multicolumn{6}{l}{\emph{SM8750}} \\
Dense & 96 & 257 & 908 & 3,226 & 11,719 \\
XAttention & 259 & 774 & 2,343 & 7,263 & 25,207 \\
ShadowAttn & 540 & 3,055 & 18,423 & 94,527 & 534,019 \\
ShadowAttn (GPU) & 712 & 2,490 & 9,051 & 33,098 & 132,945 \\
TileWeave & 96 & 226 & 603 & 1,677 & 5,470 \\
\midrule
vs.\ Dense & 1.00$\times$ & 1.14$\times$ & 1.51$\times$ & 1.92$\times$ & 2.14$\times$ \\
vs.\ XAttention & 2.70$\times$ & 3.42$\times$ & 3.89$\times$ & 4.33$\times$ & 4.61$\times$ \\
vs.\ ShadowAttn & 5.63$\times$ & 13.50$\times$ & 30.55$\times$ & 56.37$\times$ & 97.62$\times$ \\
vs.\ ShadowAttn (GPU) & 7.42$\times$ & 11.01$\times$ & 15.01$\times$ & 19.74$\times$ & 24.30$\times$ \\
\midrule
\multicolumn{6}{l}{\emph{SM8850}} \\
Dense & 70 & 190 & 614 & 1,916 & 6,879 \\
XAttention & 216 & 654 & 1,996 & 6,167 & 21,307 \\
ShadowAttn & 481 & 2,298 & 10,636 & 72,029 & 476,027 \\
ShadowAttn (GPU) & 413 & 1,510 & 5,922 & 22,505 & 89,743 \\
TileWeave & 76 & 172 & 453 & 1,193 & 3,658 \\
\midrule
vs.\ Dense & 0.92$\times$ & 1.10$\times$ & 1.35$\times$ & 1.61$\times$ & 1.88$\times$ \\
vs.\ XAttention & 2.84$\times$ & 3.80$\times$ & 4.40$\times$ & 5.17$\times$ & 5.82$\times$ \\
vs.\ ShadowAttn & 6.34$\times$ & 13.37$\times$ & 23.46$\times$ & 60.37$\times$ & 130.14$\times$ \\
vs.\ ShadowAttn (GPU) & 5.44$\times$ & 8.78$\times$ & 13.06$\times$ & 18.86$\times$ & 24.53$\times$ \\%
}

\newcommand{\selectorablationrows}{%
TileWeave & 86.37 & 57.30 \\
Keys sampled, not pooled & 85.81 & 57.31 \\
Queries pooled as well & 86.80 & 57.21 \\
XAttention's antidiagonal scores & 86.67 & 57.26 \\
Dense attention & 87.12 & 100.00 \\%
}

\newcommand{\selectortaskrows}{%
cwe & 95.78 & 100.0 & 93.22 & 51.2 & 95.08 & 51.9 & 94.48 & 57.4 \\
fwe & 64.13 & 100.0 & 64.73 & 58.9 & 64.67 & 52.1 & 64.20 & 66.1 \\
niah\_multikey\_1 & 99.80 & 100.0 & 99.80 & 54.6 & 99.20 & 55.8 & 99.80 & 57.1 \\
niah\_multikey\_2 & 85.00 & 100.0 & 83.60 & 62.0 & 84.60 & 52.3 & 84.20 & 67.4 \\
niah\_multikey\_3 & 92.20 & 100.0 & 88.60 & 55.3 & 91.00 & 50.9 & 89.00 & 60.7 \\
niah\_multiquery & 99.65 & 100.0 & 99.80 & 51.0 & 99.60 & 53.5 & 99.70 & 54.3 \\
niah\_multivalue & 90.50 & 100.0 & 87.65 & 52.6 & 90.05 & 54.5 & 88.30 & 55.6 \\
niah\_single\_1 & 100.00 & 100.0 & 100.00 & 50.7 & 100.00 & 51.6 & 100.00 & 54.1 \\
niah\_single\_2 & 100.00 & 100.0 & 100.00 & 55.7 & 100.00 & 56.6 & 100.00 & 57.7 \\
niah\_single\_3 & 99.80 & 100.0 & 99.80 & 52.9 & 99.80 & 55.1 & 99.80 & 55.3 \\
qa\_1 & 68.60 & 100.0 & 68.40 & 50.8 & 69.40 & 53.6 & 67.20 & 53.1 \\
qa\_2 & 43.00 & 100.0 & 41.60 & 44.3 & 43.00 & 52.5 & 40.80 & 47.1 \\
vt & 94.04 & 100.0 & 94.88 & 54.6 & 93.96 & 54.0 & 95.28 & 59.0 \\%
}

\newcommand{\sixteenktaskrows}{%
cwe & 1.66 & 100.0 & 2.10 & 39.2 & 2.44 & 48.5 & 2.80 & 40.6 \\
fwe & 91.67 & 100.0 & 93.53 & 44.7 & 93.07 & 54.2 & 93.40 & 45.9 \\
niah\_multikey\_1 & 96.80 & 100.0 & 93.40 & 32.9 & 94.40 & 40.1 & 92.60 & 33.9 \\
niah\_multikey\_2 & 50.40 & 100.0 & 51.80 & 45.7 & 49.20 & 53.6 & 43.00 & 46.0 \\
niah\_multikey\_3 & 37.20 & 100.0 & 32.20 & 42.3 & 34.00 & 50.9 & 30.80 & 43.3 \\
niah\_multiquery & 97.15 & 100.0 & 96.15 & 33.0 & 96.50 & 40.3 & 95.55 & 34.0 \\
niah\_multivalue & 86.20 & 100.0 & 84.20 & 32.2 & 83.05 & 39.6 & 81.35 & 33.3 \\
niah\_single\_1 & 100.00 & 100.0 & 100.00 & 38.1 & 100.00 & 44.9 & 100.00 & 37.0 \\
niah\_single\_2 & 100.00 & 100.0 & 100.00 & 32.0 & 100.00 & 39.3 & 100.00 & 33.1 \\
niah\_single\_3 & 100.00 & 100.0 & 99.80 & 33.3 & 100.00 & 40.4 & 100.00 & 34.2 \\
qa\_1 & 47.00 & 100.0 & 46.40 & 29.4 & 44.20 & 36.1 & 41.00 & 30.2 \\
qa\_2 & 34.80 & 100.0 & 33.00 & 29.9 & 34.00 & 36.5 & 31.40 & 30.4 \\
vt & 78.32 & 100.0 & 78.52 & 37.9 & 78.96 & 44.2 & 77.44 & 36.4 \\%
}

\newcommand{\llamataskrows}{%
cwe & 33.10 & 100.0 & 30.76 & 52.1 & 28.90 & 51.5 & 27.84 & 63.0 \\
fwe & 82.40 & 100.0 & 86.40 & 55.5 & 82.33 & 52.3 & 84.80 & 66.4 \\
niah\_multikey\_1 & 99.20 & 100.0 & 99.40 & 51.0 & 99.40 & 52.8 & 99.60 & 60.4 \\
niah\_multikey\_2 & 99.80 & 100.0 & 99.60 & 59.5 & 99.80 & 51.6 & 99.80 & 71.1 \\
niah\_multikey\_3 & 99.20 & 100.0 & 95.80 & 55.0 & 99.00 & 50.3 & 99.00 & 65.1 \\
niah\_multiquery & 98.35 & 100.0 & 97.65 & 52.3 & 98.00 & 53.7 & 97.70 & 61.2 \\
niah\_multivalue & 99.55 & 100.0 & 98.95 & 51.0 & 99.70 & 52.8 & 99.30 & 60.2 \\
niah\_single\_1 & 100.00 & 100.0 & 100.00 & 47.6 & 100.00 & 51.7 & 100.00 & 56.7 \\
niah\_single\_2 & 100.00 & 100.0 & 100.00 & 57.0 & 100.00 & 56.8 & 99.80 & 64.4 \\
niah\_single\_3 & 99.80 & 100.0 & 99.80 & 52.7 & 99.80 & 54.2 & 99.80 & 61.2 \\
qa\_1 & 78.40 & 100.0 & 79.60 & 53.7 & 79.00 & 54.3 & 79.40 & 61.2 \\
qa\_2 & 52.80 & 100.0 & 51.00 & 48.8 & 55.20 & 54.4 & 50.80 & 55.3 \\
vt & 85.44 & 100.0 & 79.04 & 51.7 & 86.20 & 54.3 & 84.28 & 62.0 \\%
}

\newcommand{\shadowvariantrows}{%
One ratio, $r{=}0.486$ & 86.95 & 53.41 \\
Head ratios, $r{=}0.486$ & 86.43 & 53.40 \\
One ratio, $r{=}0.2$ & 85.43 & 27.42 \\
Head ratios, $r{=}0.2$ & 8.11 & 27.44 \\%
}

\newcommand{\densityrows}{%
RULER & TileWeave & 82.1 & 69.2 & 57.3 & 49.6 & 43.8 & 41.4 \\
 & \quad $c^*{=}2$ & 84.7 & 72.5 & 60.4 & 52.7 & 46.9 & 44.8 \\
 & \qquad group mean & 80.9 & 66.1 & 53.2 & 45.2 & 39.6 & 38.2 \\
 & XAttention & 86.3 & 68.9 & 53.4 & 43.6 & 36.2 & 34.4 \\
 & ShadowAttn & -- & -- & 53.4 & 51.2 & 49.8 & 50.1 \\
\midrule
Phone & TileWeave & 76.8 & 67.7 & 56.8 & 48.5 & 43.7 & -- \\
 & \quad $c^*{=}2$ & 84.9 & 73.3 & 60.7 & 51.6 & 46.3 & -- \\
 & \qquad group mean & 80.1 & 65.6 & 51.8 & 42.4 & -- & -- \\
 & XAttention & 73.9 & 61.4 & 49.9 & 42.1 & 36.5 & -- \\%
}

\newcommand{\longbenchrows}{%
Dense & 16.98 & 42.23 & 35.57 & 29.38 & 13.75 & 22.82 & 26.79 & 37.71 & 50.25 & 50.09 & 38.20 & 22.19 & 24.13 & 37.10 \\
XAttention & 18.13 & 42.35 & 36.15 & 28.33 & 14.73 & 22.63 & 27.06 & 37.32 & 50.24 & 48.15 & 36.19 & 22.86 & 23.47 & 36.37 \\
ShadowAttn & 18.86 & 42.36 & 35.38 & 28.53 & 14.43 & 22.73 & 27.05 & 39.39 & 51.04 & 50.20 & 38.75 & 21.89 & 24.10 & 37.56 \\
TileWeave & 17.46 & 43.36 & 35.24 & 28.27 & 13.03 & 22.18 & 26.59 & 38.36 & 48.81 & 50.41 & 37.20 & 21.03 & 23.41 & 36.54 \\%
}

\makeatletter
\renewcommand{\Notice@String}{}
\makeatother
\mlsystitlerunning{TileWeave: Heterogeneous Tiling for Sparse Prefill on Mobile SoCs}
\newcommand{\us}{\ensuremath{\,\mu\mathrm{s}}\xspace}

\newcommand{\speed}[1]{\ensuremath{#1\times}}
\newcommand{\code}[1]{\texttt{\small #1}}

\begin{document}
\setlength{\footskip}{24pt}
\fancyfoot[C]{\thepage}
\twocolumn[
\mlsystitle{\texorpdfstring{\fontsize{13}{16}\selectfont\mbox{TileWeave: Heterogeneous Tiling for Sparse Prefill on Mobile SoCs}}{TileWeave: Heterogeneous Tiling for Sparse Prefill on Mobile SoCs}}
\mlsyskeywords{On-device inference, heterogeneous computing, sparse attention, KV cache, Snapdragon}
\begin{mlsysauthorlist}
\mlsysauthor{Wentao Ni}{ucsd}\quad
\mlsysauthor{Hong Cai}{qualcomm}\quad
\mlsysauthor{Mingu Lee}{qualcomm}\quad
\mlsysauthor{Chihao Yu}{ucsd}\quad
\mlsysauthor{Zhongming Yu}{ucsd}\\[0.3em]
\mlsysauthor{Hejia Zhang}{ucsd}\quad
\mlsysauthor{Kangqi Zhang}{purdue}\quad
\mlsysauthor{Fatih Porikli}{qualcomm}\quad
\mlsysauthor{Zhijian Liu}{ucsd,inco}\quad
\mlsysauthor{Jishen Zhao}{ucsd}
\end{mlsysauthorlist}
\mlsysaffiliation{ucsd}{University of California, San Diego}
\mlsysaffiliation{qualcomm}{Qualcomm AI Research}
\mlsysaffiliation{purdue}{Purdue University}
\mlsysaffiliation{inco}{Inco AI}
\mlsyscorrespondingauthor{Jishen Zhao}{jzhao@ucsd.edu}

\begin{abstract}
Smartphone applications increasingly require large language models to process long contexts, whereas high prefill latency and memory demands constrain local inference.
Block-sparse attention reduces computation and memory traffic, but its irregular selections conflict with mobile NPUs' preference for large, regular tiles.
Efficient execution must exploit cross-query reuse while avoiding redundant computation and preserving selections needed for accuracy.
In this paper, we present TileWeave, which partitions sparse attention by this cross-query reuse: highly shared key blocks run in coarse NPU tiles, while low-reuse blocks run on the GPU only for the query blocks that selected them.
In addition, we design an NPU-native pooled-key selector that keeps selection inexpensive, and a runtime that publishes GPU partials in the order the NPU consumes them and merges them in its existing output store.
On RULER, TileWeave matches XAttention's accuracy with Qwen3-1.7B and improves it with Llama-3.2-3B-Instruct; its mean LongBench scores match XAttention as well.
On Snapdragon SM8750 and SM8850 phones, it achieves 1.35--3.23$\times$ faster full-model prefill than XAttention over 1k--16k tokens and up to 1.58$\times$ faster than dense attention.
\end{abstract}
]
\begin{NoHyper}
\printAffiliationsAndNotice{}
\end{NoHyper}

\raggedbottom
\section{Introduction}
Large language models (LLMs) support applications ranging from document understanding and question answering to personal assistants.
These applications increasingly rely on long contexts to incorporate documents, retrieved information, and extended interaction histories \citep{jiang2024minference,bai2023longbench}.
Demand for these capabilities extends to edge devices, especially smartphones, where local inference keeps personal data on the device and enables operation without a network connection \citep{xue2024powerinfer}.
Compact models such as Qwen3-1.7B \citep{yang2025qwen3}, Llama-3.2-1B/3B \citep{meta2024llama32}, and Phi-3-mini \citep{abdin2024phi3} make local deployment feasible, but long contexts remain expensive even for small models.
Processing a longer prompt increases the delay before generation begins, while its growing key/value (K/V) cache consumes more of the device's limited memory \citep{xu2024npu,kwon2023efficient}.
Dense attention compounds this pressure by computing interactions across the prompt, with quadratic work and substantial memory traffic.
Block-sparse attention is therefore an important approach to efficient long-context inference: it selects informative groups of tokens and skips the remaining attention interactions, reducing computation and the associated data movement \citep{jiang2024minference,xu2025xattention,lai2025flexprefill}.
Enabling these savings on mobile hardware is essential for making long-context applications responsive within constrained device resources.

Despite progress in sparse attention, efficient execution on mobile NPUs remains challenging.
Existing sparse prefill methods primarily target GPUs, whereas mobile NPU systems emphasize efficient dense computation and heterogeneous placement of model components \citep{xu2025xattention,xu2024npu,chen2025heteroinfer}.
An NPU obtains high throughput by processing large, regular tiles and reusing their inputs, whereas neighboring queries can select different, scattered key blocks.
Executing each small selection separately repeats data staging and limits reuse; combining selections into a large tile improves hardware utilization but computes extra interactions.
Discarding selections that do not fit the shared tile can also remove information needed for accurate outputs.
Thus, fewer selected blocks do not automatically translate into lower latency: execution must preserve useful selections while exploiting the reuse that the hardware needs.

\begin{figure*}[!t]
\centering
\includegraphics[width=\textwidth,trim=0 24 0 0,clip]{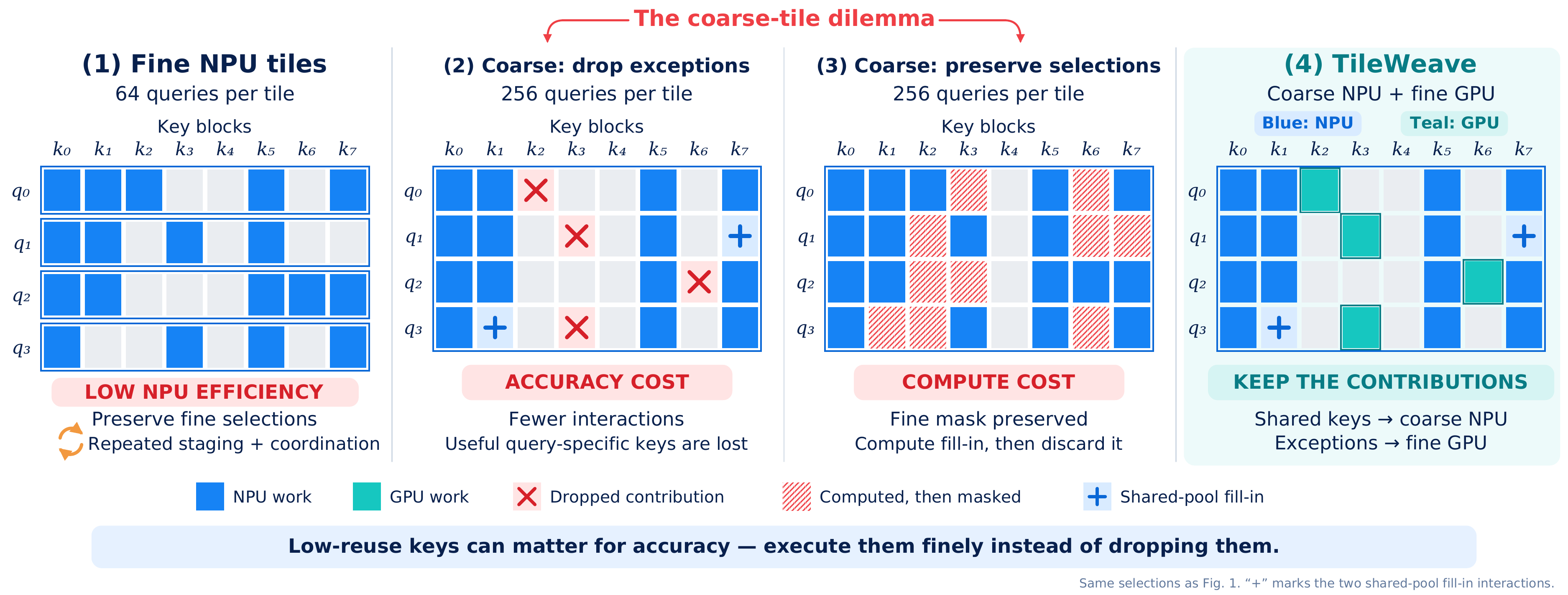}
\caption{Thin blue frames mark NPU query tiles: 64 queries in (1) and 256 in (2)--(4). Small tiles preserve selections but repeat K/V staging and coordination, limiting NPU efficiency. Coarse tiles improve reuse, but dropping exceptions risks accuracy (2), while coarse masked tiles compute unwanted interactions (3). TileWeave keeps shared keys on coarse NPU tiles and preserves exceptions through fine GPU execution (4).}
\label{fig:motivation-overview}
\end{figure*}

We introduce \emph{TileWeave}, a system that addresses such a mismatch through heterogeneous sparse prefill (Figure~\ref{fig:motivation-overview}).
Our key idea is to assign work according to how widely selected key blocks are shared across queries.
The NPU handles shared work in efficient coarse tiles, while the GPU handles query-specific work at fine granularity.
Together with inexpensive selection and coordinated execution, this split reduces wasted computation while retaining useful query-specific contributions.

In particular, our contributions span the selection policy, work partition, and runtime needed to make this approach practical.
First, we identify cross-query reuse as a criterion for choosing execution granularity, supported by measurements of NPU efficiency and the accuracy contribution of query-specific selections.
Second, we design an NPU-native selector using compact query and key summaries, coupled with a reuse-based partition that preserves every selected block while assigning it to coarse NPU or fine GPU execution.
Third, we build a runtime that overlaps the two engines and combines their partial results with low coordination overhead.
We implement TileWeave in llama.cpp \citep{gerganov2023llamacpp} and evaluate it with Qwen3-1.7B and Llama-3.2-3B-Instruct on RULER and LongBench, and on Snapdragon SM8750 and SM8850 phones.
TileWeave matches XAttention's RULER accuracy with Qwen3 and slightly exceeds it with Llama; its LongBench means remain within 0.5 points of XAttention and 0.6 points of dense attention on both models.
With Qwen3 over 1k--16k tokens, it achieves 1.35--3.23$\times$ faster full-model prefill than XAttention and up to \speed{1.58} over dense attention; attention itself is up to \speed{2.14} faster than dense attention, including selection and coordination.

\section{Background and Related Work}
\label{sec:background}
\paragraph{Attention and sparsity.}
Prefill processes prompt tokens in parallel and writes their keys and values to the K/V cache; decode generates one token at a time \citep{kwon2023efficient}.
Block-sparse prefill selects pairs of query and key blocks; attention density is the retained fraction of causal query-key pairs.
Grouped-query attention (GQA) shares each K/V head among several query heads, allowing the kernel to reuse loaded K/V rows \citep{ainslie2023gqa}.
FlashAttention computes tiled attention using online-softmax statistics, which can be combined across disjoint key subsets to normalize their outputs jointly \citep{dao2022flashattention,milakov2018online}.

\begin{figure}[!t]
\centering
\includegraphics[width=\columnwidth]{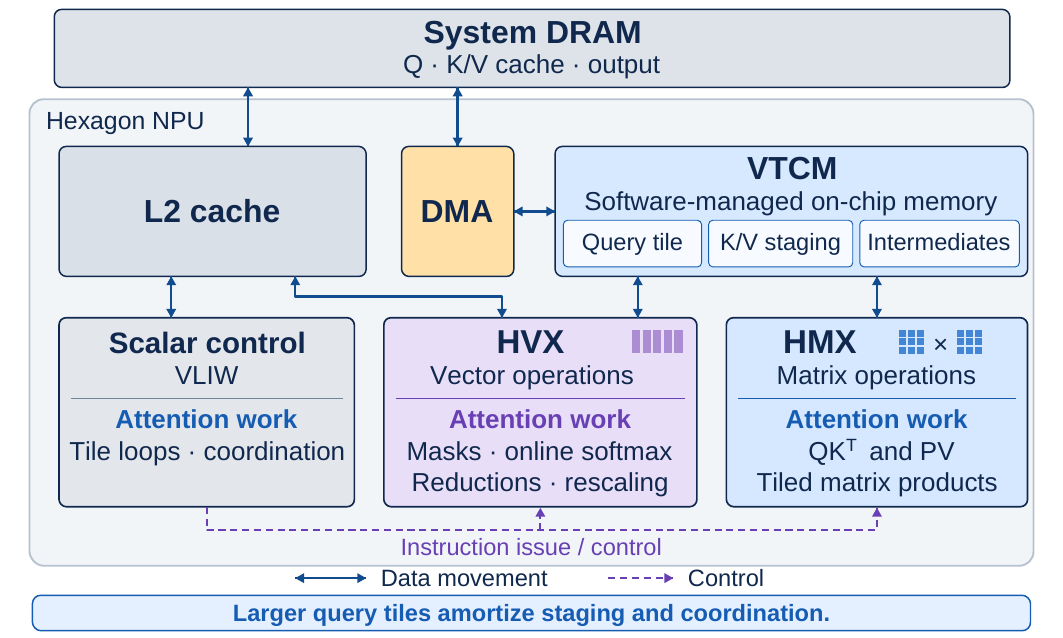}
\caption{Hexagon NPU architecture and attention mapping. Larger query tiles amortize K/V staging and coordination.}
\label{fig:npu-architecture}
\end{figure}

\paragraph{Phone NPU architecture.}
Snapdragon SoCs integrate a Hexagon NPU and an Adreno GPU sharing DRAM.
Figure~\ref{fig:npu-architecture} shows Hexagon's scalar control threads, Hexagon Vector eXtensions (HVX), and Hexagon Matrix eXtensions (HMX) \citep{mahurin2023hexagon}.
In attention, HMX executes matrix products, while HVX handles masks, reductions, and softmax.
FP16 HMX operates on $32\times32$ hardware tiles in a dedicated memory layout \citep{hao2026scaling}.
These are building blocks for larger query tiles; sparse selection uses its own granularity.
HMX operands must be staged from DRAM into software-managed on-chip tightly coupled memory (VTCM), whose buffers are reused as computation advances.
Direct memory access (DMA) transfers can overlap data movement with computation \citep{mahurin2023hexagon}.
Coarse query tiles reuse each staged K/V block across more queries and amortize staging and synchronization; fine tiles repeat these steps for smaller amounts of useful work.
The GPU supports indexed loads for traversing selected blocks, while the NPU favors large, regular matrix operations \citep{chen2025heteroinfer}.
Shared allocations avoid copying the K/V cache between engines, but their separate caches and shared memory bandwidth require explicit publication and coordination.

\paragraph{Sparse attention methods.}
For prefill, MInference assigns sparse patterns to heads offline and constructs their indices at run time \citep{jiang2024minference}; XAttention estimates block importance from strided antidiagonal scores \citep{xu2025xattention}.
SpargeAttn predicts sparsity from compressed query/key blocks \citep{zhang2025spargeattn}; FlexPrefill probes recent queries \citep{lai2025flexprefill}, while SampleAttention samples query chunks across the sequence \citep{zhu2024sampleattention}.
FlashPrefill scores pooled keys and selects blocks relative to each row's maximum \citep{fan2026flashprefill}.
Learned gates and architectures such as SeerAttention, NSA, and MoBA incorporate sparse selection during training \citep{gao2024seerattention,yuan2025native,lu2025moba}.
For decoding, sparse attention reduces K/V-cache reads through query-aware retrieval or adaptive token selection \citep{tang2024quest,liu2025clusterkv,lin2025twilight,ni2026doublep}.

\paragraph{Heterogeneous mobile inference.}
Prior NPU inference systems use prompt chunking and heterogeneous scheduling \citep{xu2024npu}; PowerInfer-2 distributes neuron clusters across CPU and NPU \citep{xue2024powerinfer}.
HeteroInfer combines layer-level placement with tensor-level GPU/NPU parallelism, using profiled performance to choose weight, activation, or hybrid partitions \citep{chen2025heteroinfer}.
ShadowAttn, part of ShadowNPU, estimates token importance with dense INT8 scores on the NPU, then selects keys per query row and evaluates their attention on the CPU or GPU, overlapping stages across heads \citep{yin2026shadownpu}.

\section{Motivation}
\label{sec:motivation}
The cost of sparse attention on an NPU depends on how selected key blocks are shared across queries.
Large query tiles improve execution efficiency, but neighboring query blocks can select different keys.
Executing their union either introduces additional attention pairs or computes contributions that masking later discards.
Cross-query reuse identifies where finer execution can reduce this excess work.

\subsection{Coarse Execution Couples Query Selections}
\label{sec:motivation-tiles}
The query selection block $b_q$, key selection block $b_k$, and NPU query tile $B_r$ serve different purposes.
The first two determine which query-key pairs can be selected together; the last determines how the kernel shares staging and computation.
FlashAttention-style prefill keeps $B_r$ queries resident while DMA stages K/V chunks into VTCM, HMX computes matrix products, and HVX applies masks and online softmax.
A larger query tile amortizes staging and pipeline coordination over more queries.

\paragraph{Coarse tiles improve NPU efficiency.}
For $N_q$ queries, $H_q$ query heads, head dimension $D$, and $S$ selected keys per query, nominal matrix work is $4H_qN_qSD$, independent of $B_r$.
At a fixed K/V chunk width $C$, however, the kernel visits $(N_q/B_r)\lceil S/C\rceil$ chunks per K/V head.
Reducing $B_r$ from 256 to 64 quadruples these visits.
In the equal-work experiment of Table~\ref{tab:motivation}, with $S=2048$, 64-, 128-, and 256-query tiles take 4.121, 3.277, and 2.344 ms.
The largest tile is \speed{1.76} faster at the same selected work.

\begin{table}[!tb]
\caption{NPU attention latency in ms (speedup over $B_r=64$). Larger query tiles improve efficiency, giving up to \speed{1.76} speedup at fixed selected work.}
\label{tab:motivation}
\vspace{0.1in}
\centering
\small
\setlength{\tabcolsep}{4pt}
\begin{tabular}{lrrr}
\toprule
Query tile $B_r$ & $S=1024$ & $S=1536$ & $S=2048$ \\
\midrule
64  & 2.62 (\speed{1.00}) & 3.18 (\speed{1.00}) & 4.12 (\speed{1.00}) \\
128 & 2.15 (\speed{1.22}) & 2.53 (\speed{1.25}) & 3.28 (\speed{1.26}) \\
256 & 1.51 (\speed{1.73}) & 1.82 (\speed{1.74}) & 2.34 (\speed{1.76}) \\
\bottomrule
\end{tabular}
\end{table}

\paragraph{Sharing selections changes work or attention.}
Serving four adjacent 64-query selections from one 256-query tile requires the tile to traverse their union of key blocks.
Masking the tile back to each query block's selection preserves its attended pairs, but the matrix products still compute the unwanted interactions.
An unmasked union includes those additional keys in attention; under a fixed density budget, this fill-in also competes with query-specific selections.

\paragraph{Low reuse does not imply low importance.}
Dropping rarely shared selections reduces work but can remove useful contributions.
Table~\ref{tab:motivation-sharing} holds the final selector and threshold fixed on Llama-3.2-3B-Instruct over 6,500 RULER 4k prompts.
The pool retains blocks requested by at least three of four query blocks; the pool-plus-diagonal control removes only non-diagonal exceptions, preserving sink blocks, head-start tiles, and dense tails.
Restoring those query-specific exceptions raises accuracy from 85.57 to 86.32, a \motivationqualitygain-point gain (paired 95\% bootstrap interval: [\motivationqualitycilow, \motivationqualitycihigh]).
The union obtains a similar score but retains 73.13\% of causal pairs, against 62.19\% with exceptions.
These are fixed-configuration policies, not density-matched selectors; each recomputes its pools from its own activations.
The result supports preserving low-reuse contributions while reducing their excess coarse work.

\begin{table}[!tb]
\caption{Llama-3.2-3B on RULER 4k. Non-diagonal exceptions add \motivationqualitygain\ points; the union retains more pairs for a similar score.}
\label{tab:motivation-sharing}
\vspace{0.1in}
\centering
\small
\begin{tabular}{lrr}
\toprule
Attention policy & Score & Density (\%) \\
\midrule
\motivationqualityrows
\bottomrule
\end{tabular}
\end{table}

\paragraph{Selection axes have different effects.}
Coarsening selection itself is another option, whose quality depends on which axis changes.
At matched density on Qwen3 RULER 4k, XAttention scores 86.39 with $b_q=b_k=64$ and 86.29 with $b_q=256$, $b_k=64$; coarsening both axes to 256 lowers the score to 84.00.
This control supports retaining fine key resolution, but does not establish a universal need for fine query blocks.
Our objective is to execute query-specific selections efficiently where they are useful, while retaining coarse execution where selections overlap.

\subsection{Cross-Query Reuse Determines Excess Work}
\label{sec:motivation-reuse}
The extra computation of a coarse tile is unevenly distributed across its key blocks.
For one K/V head, let $W_0,\ldots,W_3$ be the selections of four adjacent 64-query blocks, $U$ their union, and $c(b)$ the number of query blocks selecting key block $b$.
For fully visible blocks, coarse execution performs four $64\times64$ interactions for every $b\in U$, while the fine selections request only $c(b)$.
The number of interactions beyond those selections is
\begin{equation}
\Delta W=\sum_{b\in U}\bigl(4-c(b)\bigr).
\label{eq:reuse-excess}
\end{equation}
A fully shared block uses all four coarse interactions; a singly requested block creates three additional interactions.
Fine execution replaces four coarse NPU interactions with four requested interactions for the shared block, but only one for the singly requested block.

Cross-query reuse thus assigns execution granularity: high-reuse blocks retain coarse execution, while low-reuse blocks save more coarse work per requested interaction moved.
Their quality contribution persists with diagonal attention preserved (Table~\ref{tab:motivation-sharing}), motivating fine execution that retains these contributions while reducing excess coarse work.

\subsection{Reuse Partitioning Creates an Execution Opportunity}
\label{sec:motivation-offload}
The mobile GPU provides a second execution path for blocks whose low reuse makes coarse NPU execution expensive.
We quantify the opportunity on SM8750 with Llama-3.2-3B-Instruct, timing NPU attention with fixed Q/K/V tensors and group-mean selections, then removing whole key blocks selected by at most $r$ of the four query blocks.
Let $T_{\mathrm{all}}$ be the original attention time and $T_{\mathrm{remain}}(r)$ the measured time for the retained blocks.
If removed contributions complete at zero execution, transfer, contention, and merge cost, the ideal speedup is
\begin{equation}
S_{\mathrm{ideal}}(r)=\frac{T_{\mathrm{all}}}{T_{\mathrm{remain}}(r)}.
\label{eq:offload-bound}
\end{equation}
The bound holds the NPU kernel, inputs, and tile geometry fixed; the complete attention result still requires the removed contributions.

\begin{table}[!tb]
\caption{Ideal NPU speedup for Llama-3.2-3B assuming removed blocks complete for free. Offloading blocks selected by at most two query blocks gives \speed{\offloadtwospeedfour}--\speed{\offloadtwospeedeight} speedup.}
\label{tab:motivation-offload}
\vspace{0.1in}
\centering
\small
\begin{tabular}{lrr}
\toprule
& \multicolumn{2}{c}{Ideal speedup} \\
\cmidrule(lr){2-3}
Selection count offloaded & 4k & 8k \\
\midrule
None (baseline) & \speed{1.00} & \speed{1.00} \\
1 of 4 & \speed{\offloadonespeedfour} & \speed{\offloadonespeedeight} \\
At most 2 of 4 & \speed{\offloadtwospeedfour} & \speed{\offloadtwospeedeight} \\
At most 3 of 4 & \speed{\offloadthreespeedfour} & \speed{\offloadthreespeedeight} \\
\bottomrule
\end{tabular}
\end{table}

In this fixed-input study, singly selected blocks account for about 8\% of requested block interactions but 22\% of coarse NPU interactions, counted before token-level causal masking.
Removing blocks selected by at most two query blocks yields ideal NPU speedups of \speed{\offloadtwospeedfour} at 4k and \speed{\offloadtwospeedeight} at 8k (Table~\ref{tab:motivation-offload}).
The measurements cover all 28 layers and all sparse chunks of the first variable-tracking prompt at each length; they exclude selection, host dispatch, dense tails, and every cost of completing the removed work.

Realizing this opportunity requires inexpensive selection and a second execution path whose computation, publication, and merge costs remain below the NPU work saved.
TileWeave addresses these requirements with an NPU-native selector, a reuse-based partition, and coordinated execution within each attention operator (Section~\ref{sec:method}).

\section{Selecting Finely, Executing Coarsely}
\label{sec:method}
\begin{figure*}[!t]
\centering
\includegraphics[width=\textwidth]{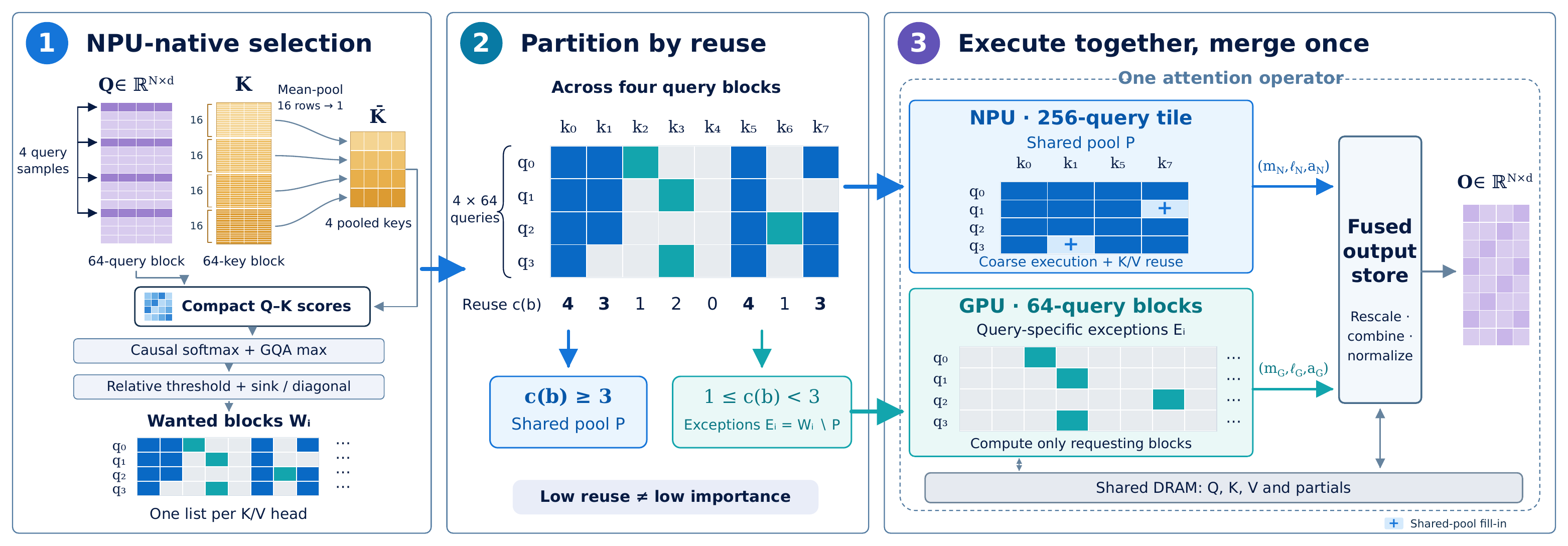}
\caption{TileWeave overview. (1) Four sampled query rows and four pooled key vectors per 64-token block produce compact scores; causal softmax, GQA maximum, and relative thresholding with forced sink/diagonal blocks produce wanted lists $W_i$ per K/V head. (2) With $c^*=3$, reuse across four query blocks assigns key blocks to the shared NPU pool $P$ ($c(b)\geq3$, blue) or query-specific GPU exceptions $E_i=W_i\setminus P$ ($1\leq c(b)<3$, teal). The masks illustrate the partition; plus signs mark shared-pool fill-in. (3) Coarse NPU tiles and fine GPU blocks execute together, and a fused output store rescales, combines, and normalizes their online-softmax partials into $O$.}
\label{fig:design}
\end{figure*}

Section~\ref{sec:motivation} motivates preserving useful query-specific contributions while reducing their excess coarse work.
TileWeave selects at 64-query granularity and assigns execution by reuse: with $c^*=3$, keys requested by three or four query blocks run in a coarse 256-query NPU tile; the rest run on the GPU only for requesting blocks.
Figure~\ref{fig:design} outlines inexpensive selection (Section~\ref{sec:selection}), reuse-based partitioning (Section~\ref{sec:prefill}), and coordinated execution (Section~\ref{sec:contract}); Section~\ref{sec:ablations} evaluates the alternatives.

\subsection{Fine Selection at NPU Cost}
\label{sec:selection}
\label{sec:motivation-selectors}
\begin{figure}[!tb]
\centering
\includegraphics[width=\columnwidth]{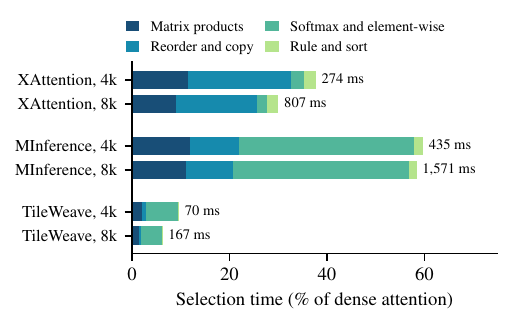}
\caption{Qwen3-1.7B NPU selection cost relative to dense attention (totals in ms). TileWeave reduces selection to 6--10\% of dense attention's time.}
\label{fig:selector-breakdown}
\end{figure}

\paragraph{Keep selection within the execution budget.}
GPU-designed selectors use reorders, copies, and reductions that run as separate NPU vector operators, moving operands between DRAM and VTCM (Section~\ref{sec:background}).
Profiling Qwen3-1.7B on one RULER prompt per length, XAttention spends 154 of its 274\,ms of selection at 4k on SM8750 reordering queries and copying keys, against 83\,ms on matrix products.
MInference's vertical-slash estimate spends 262 of 435\,ms on softmax and element-wise operators (Figure~\ref{fig:selector-breakdown}).
Their per-query-head lists also require separate execution lanes or a denser union: the GQA kernel stages K/V once per group and executes one list per K/V head.
Our selector uses compact matrix products and emits that shared list directly.

\paragraph{Compact representatives.}
One compact matrix product scores 64-query/64-key block pairs using four sampled queries and four pooled keys.
For query block $i$, the samples are rows $0,16,32,48$ of every query head; for K/V block $j$, each pool is the mean of 16 consecutive keys.
For K/V head $h$, GQA group size $G$, query-head offset $g$, and sample and pool indices $u,v\in\{0,1,2,3\}$, define
\begin{equation}
\begin{aligned}
Q_{ihgu}&=q_{64i+16u,hG+g},\\
K_{jhv}&=\tfrac1{16}\textstyle\sum_{t=0}^{15}k_{64j+16v+t,h},\\
z_{ijhguv}&=Q_{ihgu}^{\top}K_{jhv}/\sqrt D.
\end{aligned}
\label{eq:representatives}
\end{equation}
Q indices are local to the current prefill chunk; K indices address the visible cache.

\paragraph{Pool keys, sample queries.}
By linearity, $z_{ijhguv}$ is the mean logit over all 16 keys in pool $v$, so every key contributes; sampling can miss a block's few target tokens.
Neighboring queries attend to similar keys, so four samples retain most of their ranking; pooling queries gains under half a point at more NPU time (Section~\ref{sec:selector-ablation}).
A strided Q view and a sixteenth of K avoid query reordering and key copying.
Each score costs one $D$-dimensional product, versus $16D$ for XAttention's antidiagonals over the same score-matrix size.
Key pools reside beside the K/V cache; each prefill chunk pools only newly written blocks.

\paragraph{Normalize, then take each group's maximum.}
Averaging query samples or heads before scoring can collapse distinct query directions.
Instead, we normalize each sample and head separately over the key pools it can see, then aggregate their probabilities.
Sample $u$ of query block $i$ sees every pool of earlier blocks and, inside block $i$, the pools $w\le u$ that start no later than it does; with $V_{iu}=\{(b,w):b<i\}\cup\{(i,w):w\le u\}$,
\begin{equation}
\begin{aligned}
a_{ijhguv}&=\frac{\exp(z_{ijhguv})}{\sum_{(b,w)\in V_{iu}}\exp(z_{ibhguw})},\\
p_{ijh}&=\max_{0\le g<G}\frac14\sum_{u=0}^{3}\sum_{v:(j,v)\in V_{iu}}a_{ijhguv}.
\end{aligned}
\label{eq:representative-probability}
\end{equation}
Summing visible pool probabilities and averaging query samples gives each query head's block score.
Taking the maximum merges their selections: a block enters the K/V head's list whenever any query head selects it.
Averaging the heads instead halves ($G=2$) the score of a block that only one head needs, which at 16k drops retrieval blocks (Appendix~\ref{app:head-sharing}).
$A_i=\{0,\ldots,i\}$ is the set of blocks block $i$ can reach; the attention executor later enforces exact token causality.

\paragraph{Select the wanted blocks.}
We compare each probability with the average probability over reachable blocks:
\begin{equation}
W_{i,h}=\{j\in A_i:p_{ijh}|A_i|>\tau\}\cup\{0,d_i\},
\label{eq:membership}
\end{equation}
where $d_i$ is the chunk-offset diagonal block and block zero is the sink.
The threshold lets the number of selected blocks follow the estimated distribution, and one $\tau$ serves every layer, head, and length.
Sink and diagonal membership are forced after thresholding so that they do not distort the other probabilities.

\subsection{Execution Granularity from Cross-Query Reuse}
\label{sec:prefill}
Cross-query reuse determines a selected block's excess coarse work (Section~\ref{sec:motivation-reuse}).
For one K/V head, suppress $h$ and count requests across four adjacent wanted lists $W_0,\ldots,W_3$:
\begin{equation}
\begin{aligned}
U &= \bigcup_{i=0}^{3}W_i,\qquad c(b)=\sum_{i=0}^{3}\mathbf{1}[b\in W_i],\\
P &= \{b\in U:c(b)\geq c^*\},\\
E_i &= W_i\setminus P.
\end{aligned}
\label{eq:partition}
\end{equation}
Assignments follow each tile's current wanted lists, even at fixed shapes and $c^*$.
The NPU evaluates $P$ for all 256 queries, retaining coarse tiles and K/V reuse; the GPU evaluates $E_i$ only for its requesting 64-query block.
At $c^*=3$, exceptions requested by one or two query blocks require one or two GPU interactions before causal masking.
Setting $c^*=1$ executes the union $U$ entirely on the NPU; larger thresholds reduce pool fill-in and move more selections to the GPU.
The NPU-only masked tile executes $U$ but retains only each $W_i$, computing and discarding fill-in.
We use $c^*=3$; Section~\ref{sec:executions} compares these executions at a fixed selector and threshold.

The effective key set is $P\cup E_i$, with disjoint partitions and $W_i\subseteq P\cup E_i\subseteq U$.
Every wanted block is retained, but shared blocks can add neighboring queries' selections, unlike the naive 64-block baseline's direct execution of $W_i$.

\subsection{One Operator on Two Engines}
\label{sec:contract}
\begin{figure*}[!t]
\centering
\includegraphics[width=\textwidth]{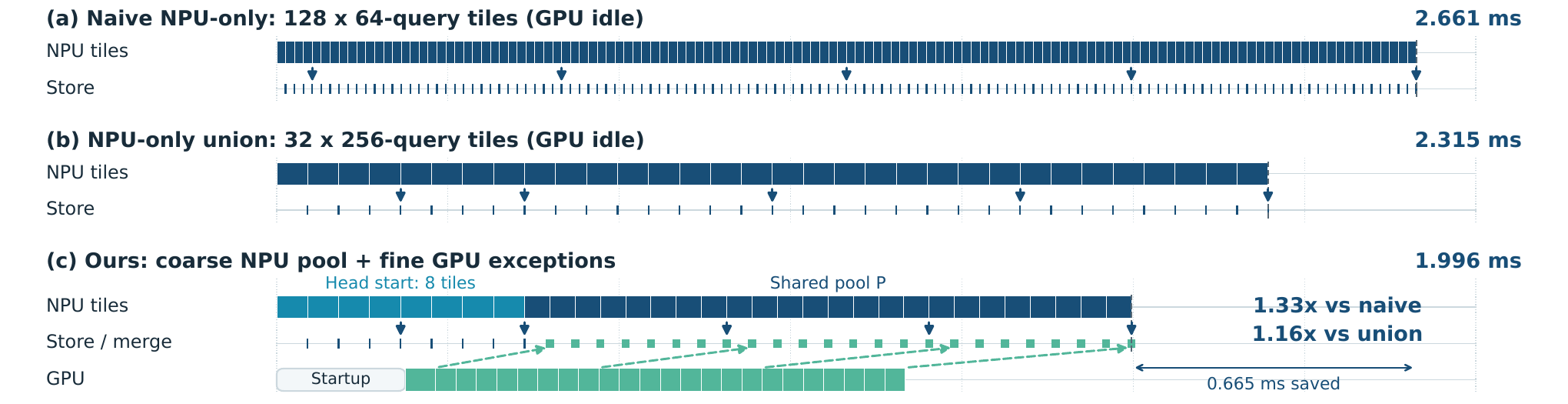}
\caption{Fine-grained schedules on a shared time axis (1024 queries, eight K/V heads). Endpoints are measured two-round means from the historical 4k Qwen3-1.7B/SM8750 study using the earlier threshold scorer; selection finishes before $t=0$ and other model operators are excluded. Tile intervals are approximate; store/merge marks encode no duration. GPU completion spans 0.37--1.47\,ms from a nearby trace, with intermediate groups interpolated. Timing calibration illustrates overlap and relative latency, not a single event trace.}
\label{fig:pipeline}
\end{figure*}

The NPU and GPU compute disjoint key subsets within one layer's attention over a prefill chunk.
The split pays off when GPU execution, coordination, and merging cost less than the coarse work removed.
Figure~\ref{fig:pipeline} compares the three schedules using historical timing calibration (Appendix~\ref{app:runtime-diagnostics}).
For a query and selected key subset $S$, an engine emits
\begin{equation}
\begin{aligned}
m_S&=\max_{j\in S}s_j,\qquad \ell_S=\sum_{j\in S}e^{s_j-m_S},\\
a_S&=\sum_{j\in S}e^{s_j-m_S}v_j.
\end{aligned}
\end{equation}
Let $m=\max(m_N,m_G)$ and $w_e=e^{m_e-m}$ for engine $e$.
The output is
\begin{equation}
O=\frac{w_Na_N+w_Ga_G}{w_N\ell_N+w_G\ell_G}.
\label{eq:merge}
\end{equation}
An empty subset contributes zero weight and accumulator.
This standard online-softmax composition \citep{milakov2018online,dao2022flashattention} is algebraically exact for the effective selected set, with numerical differences determined by the two implementations; it does not make the sparse policy equivalent to dense attention.
The runtime overlaps the engines, merges in the existing store, and publishes shared data explicitly; Appendix~\ref{app:runtime-diagnostics} records the supporting timing and correctness diagnostics.

\paragraph{Overlap startup and publish in consumption order.}
GPU work depends on the NPU's selections.
The first eight execution tiles therefore keep the union $U$ on the NPU to cover GPU startup; later tiles use Equation~\ref{eq:partition}.
The accuracy runs include these tiles' union fill-in in the effective mask.
GPU work follows the NPU's store order in query-block/K/V-head stages: persistent workgroups process count-bounded exception queues and publish each stage when its last workgroup finishes.
The NPU waits only for the stage its current store needs, avoiding a whole-operator GPU wait or a separate launch per stage.

\paragraph{Merge in the existing store.}
HTP store workers combine the GPU's statistics with the raw HMX accumulator and normalize once, avoiding a separate pass over the NPU output.
They handle HMX's transformed base-2 maximum, and both engines apply the same causal mask.
The GPU writes F16 partials in NPU tile-row order for vectorized merging; tile-ahead prefetches bring completed rows in before the store needs them.

\paragraph{Make each publication visible.}
The GPU imports shared rpcmem allocations through OpenCL without copying the K/V cache, using kernel layouts and an import path derived from the local mllm research fork \citep{mllm}.
An atomic ready word alone does not make new payloads visible through the engines' separate caches.
After observing it, a host relay submits a GPU operation referencing the newly written Q and membership buffers before opening the GPU gate.
NPU readers invalidate published rows unless a whole-L2 flush already did so, and prefetch exact row ranges to avoid caching future partials.

\paragraph{Bound the GPU's share.}
Both engines share DRAM bandwidth, so excess GPU work can delay the NPU.
The reuse threshold limits offload to low-reuse blocks; fully shared selections leave no GPU work.
An optional per-tile budget returns excess exceptions to the NPU, masked to their requesting query blocks so the effective mask is unchanged.
Incomplete prefill tails use dense attention.

\section{Evaluation}
\label{sec:evaluation}
We first compare TileWeave with dense attention, XAttention, and ShadowAttn in accuracy, on RULER and LongBench with two models (Section~\ref{sec:accuracy}), and in performance, as attention and full-model prefill latency on two phones from 1k to 16k tokens (Section~\ref{sec:performance}); we then ablate the selector and its execution, including NPU-only executions with the group-maximum selector and threshold held fixed (Section~\ref{sec:ablations}).

\subsection{Experimental Setup}
\label{sec:setup}
\paragraph{Models and platforms.}
We evaluate Qwen3-1.7B \citep{yang2025qwen3} and Llama-3.2-3B-Instruct \citep{meta2024llama32}: 28 layers, eight K/V heads, and head dimension 128, with 16 and 24 query heads.
Both use Q4\_0 GGUF weights (Qwen3 with a Q6\_K output projection, Llama with a tied Q6\_K embedding) and F16 K/V.
Accuracy uses CUDA reference implementations with these weights dequantized to F16 and dense greedy decode in every method, so only prefill attention differs.
Performance runs use Qwen3-1.7B, and Llama-3.2-3B-Instruct for the comparison of executions, in llama.cpp \citep{gerganov2023llamacpp} on Snapdragon SM8750 phones (Hexagon HTPv79 NPU, Adreno 830 GPU), and Qwen3-1.7B's latency also on SM8850 (HTPv81, Adreno 840).

\paragraph{Methods.}
We compare dense attention with two training-free sparse methods.
\emph{XAttention} \citep{xu2025xattention} keeps, for each query block and query head, the 64-token key blocks with the highest antidiagonal scores up to a cumulative threshold; on the phone it runs on our block-sparse NPU kernel.
Our \emph{ShadowAttn} reimplementation \citep{yin2026shadownpu} keeps a fixed ratio of each query row's keys and evaluates NPU--CPU and NPU--GPU execution.
The CPU version gathers selected keys; the GPU version applies the per-row selections as a FlashAttention mask.
Phone estimation uses F16 HMX products mapped to 8-bit scores, and estimation and attention run sequentially, without the published head-wise pipeline (Appendix~\ref{app:shadowattn-execution}).
\emph{TileWeave} is pooled-key selection with the NPU-GPU split at $c^*=3$, with the tile-row merge and GPU kernel of Section~\ref{sec:contract}; the \emph{union} and the \emph{masked tile} run the same selector on the NPU alone (Section~\ref{sec:prefill}).
Each sparse method has one setting per model, kept at every length and benchmark: XAttention's threshold, ShadowAttn's ratio, and our $\tau$ are 0.9268, 0.486, and 0.4277 for Qwen3-1.7B, and 0.8736, 0.488, and 0.5582 for Llama-3.2-3B-Instruct.

\paragraph{Benchmarks.}
RULER \citep{hsieh2024ruler} has 13 tasks with 500 prompts each, at 4k, 8k, and 16k tokens (Qwen3-1.7B also at 1k and 2k, Appendix~\ref{app:selector-tasks}); we report the equal-task mean score.
LongBench v1 \citep{bai2023longbench} adds single-document QA (Qasper, MultiFieldQA-en/MFQA), multi-document QA (HotpotQA/HQA, 2WikiMultihopQA/2WQA, MuSiQue/MSQ), and query-based summarization (QMSum), 200 prompts each (150 for MultiFieldQA-en) truncated to 16k tokens, with the official metrics.

\paragraph{Measurement protocol.}
Phone prefill latency uses five RULER variable-tracking prompts per length, with one warmup followed by two timed prefills per prompt from an empty cache; ShadowAttn uses a 1k-token warmup and one timed prefill per prompt.
We run two or three interleaved rounds and report the median across rounds of the mean per-prompt median latency.
Attention latency directly times each chunk's attention on captured inputs, including selection and coordination, with one warmup prompt and one timed run per chunk (ShadowAttn CPU: two prompts at 16k).
Attention and full-model prefill are measured in separate runs (Appendix~\ref{app:attention-timing}).
All phone runs use two CPU workers pinned to the two fastest cores (6--7, affinity mask \code{0xc0}).

\subsection{Accuracy}
\label{sec:accuracy}
\begin{table}[!tb]
\caption{RULER mean accuracy. TileWeave matches XAttention on Qwen3-1.7B and is slightly more accurate on Llama-3.2-3B-Instruct at every tested length.}
\label{tab:ruler}
\vspace{0.1in}
\centering
\small
\setlength{\tabcolsep}{4pt}
\begin{tabular}{lrrrrrr}
\toprule
& \multicolumn{3}{c}{Qwen3-1.7B}
& \multicolumn{3}{c}{Llama-3.2-3B-Instruct} \\
\cmidrule(lr){2-4}\cmidrule(lr){5-7}
Method & 4k & 8k & 16k & 4k & 8k & 16k \\
\midrule
Dense        & 87.12 & 79.48 & 70.86 & 86.77 & 83.39 & 80.61 \\
XAttention   & 86.31 & 77.16 & 70.08 & 86.00 & 82.64 & 80.13 \\
ShadowAttn   & 86.95 & 78.51 & 69.54 & 86.72 & 83.49 & 80.71 \\
TileWeave & 86.37 & 77.19 & 69.99 & 86.32 & 82.98 & 80.19 \\
\bottomrule
\end{tabular}
\end{table}

\begin{table*}[!t]
\caption{LongBench accuracy (F1; ROUGE-L for QMSum). TileWeave's mean stays within 0.6 points of dense attention on both models.}
\label{tab:longbench}
\vspace{0.1in}
\centering
\footnotesize
\setlength{\tabcolsep}{3pt}
\begin{tabular}{lrrrrrrrrrrrrrr}
\toprule
& \multicolumn{7}{c}{Qwen3-1.7B} & \multicolumn{7}{c}{Llama-3.2-3B-Instruct} \\
\cmidrule(lr){2-8}\cmidrule(lr){9-15}
Method & Qasper & MFQA & HQA & 2WQA & MSQ & QMSum & Avg. & Qasper & MFQA & HQA & 2WQA & MSQ & QMSum & Avg. \\
\midrule
\longbenchrows
\bottomrule
\end{tabular}
\end{table*}

\paragraph{RULER.}
TileWeave matches XAttention with Qwen3-1.7B from 1k to 16k tokens and slightly exceeds it with Llama-3.2-3B-Instruct at every tested length (Table~\ref{tab:ruler}, Appendix~\ref{app:selector-tasks}).
Relative to dense attention, its mean scores decrease by at most 2.3 points with Qwen3 and 0.5 points with Llama.
ShadowAttn's per-row selection generally achieves higher accuracy, with a notable advantage on Qwen3's common-words extraction task, where both block-sparse methods lose accuracy.
At 16k, however, TileWeave slightly exceeds ShadowAttn with Qwen3.
The group-maximum rule's contribution to long-context retrieval is examined in Appendix~\ref{app:head-sharing}.

\paragraph{LongBench.}
Across the question-answering and summarization tasks, TileWeave's mean scores remain within 0.5 points of XAttention and 0.6 points of dense attention with both models (Table~\ref{tab:longbench}).
It slightly trails XAttention with Qwen3 and slightly exceeds it with Llama, while ShadowAttn achieves higher means with both models.
These comparisons use the same method settings as RULER, without retuning for individual datasets.

\subsection{Performance}
\label{sec:performance}
\begin{figure*}[!t]
\centering
\includegraphics[width=\textwidth]{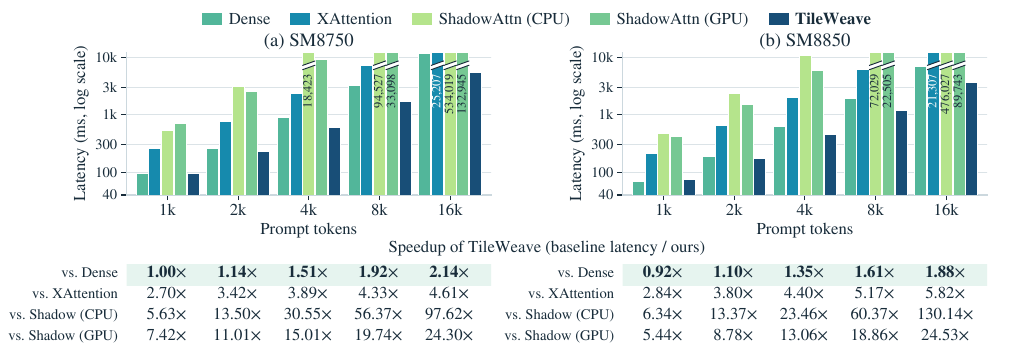}
\caption{Qwen3-1.7B attention latency, including selection and coordination. TileWeave achieves speedups of up to \speed{2.14} over dense attention, \speed{5.82} over XAttention, \speed{130.14} over ShadowAttn (CPU), and \speed{24.53} over ShadowAttn (GPU).}
\label{fig:attention-latency}
\end{figure*}

\begin{figure*}[!t]
\centering
\includegraphics[width=\textwidth]{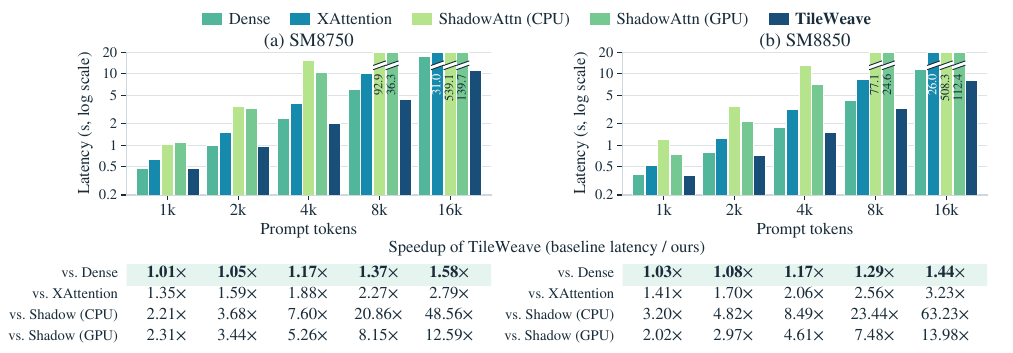}
\caption{Qwen3-1.7B full-model prefill latency over 1k--16k tokens. TileWeave achieves speedups of up to \speed{1.58} over dense attention, \speed{3.23} over XAttention, \speed{63.23} over ShadowAttn (CPU), and \speed{13.98} over ShadowAttn (GPU).}
\label{fig:prefill-latency}
\end{figure*}

\paragraph{Prefill.}
With Qwen3-1.7B on SM8750, TileWeave matches dense attention at 1k tokens and is faster from 2k on: \speed{1.05}, \speed{1.17}, \speed{1.37}, and \speed{1.58} at 2k, 4k, 8k, and 16k (Figure~\ref{fig:prefill-latency}).
The gain grows with length, consistent with the quadratic growth of attention work.
XAttention is 1.33--1.77$\times$ slower than dense attention at every length, so TileWeave is 1.35--2.79$\times$ faster than it, and the gap widens with length.
ShadowAttn's CPU version is slower still: its prefill takes 1.04\,s at 1k and 539\,s at 16k, 2.2--30.8$\times$ dense attention's, so TileWeave is 2.2--48.6$\times$ faster than it.
On SM8850 TileWeave is \speed{1.03}, \speed{1.08}, \speed{1.17}, \speed{1.29}, and \speed{1.44} faster than dense attention from 1k to 16k, 1.41--3.23$\times$ faster than XAttention, and 3.2--63.2$\times$ faster than ShadowAttn's CPU version.
The GPU version is faster than the CPU version from 2k onward, but TileWeave remains 2.3--12.6$\times$ faster on SM8750 and 2.0--14.0$\times$ on SM8850.

\paragraph{Attention.}
Figure~\ref{fig:attention-latency} reports directly measured attention latency.
At 1k its selection costs about what the sparse kernels save, so TileWeave's attention matches dense attention; at 2k, 4k, 8k, and 16k it is \speed{1.14}, \speed{1.51}, \speed{1.92}, and \speed{2.14} faster.
On SM8850 it takes 9\% longer at 1k and is \speed{1.10}, \speed{1.35}, \speed{1.61}, and \speed{1.88} faster from 2k to 16k.
XAttention's attention is 2.2--3.0$\times$ slower than dense attention on SM8750 and 3.1--3.4$\times$ on SM8850, although it executes fewer blocks than TileWeave (Appendix Table~\ref{tab:density}).
In a profiled prefill its selection is only 12--18\% of that time and about three times TileWeave's; the rest runs its per-head masks as 64-query tiles in two GQA lanes, the small tiles that Section~\ref{sec:motivation-tiles} shows are inefficient on the NPU.
ShadowAttn's CPU attention takes 5.7--46$\times$ as long as dense attention and 5.6--98$\times$ as long as TileWeave's on SM8750, and 6.9--69$\times$ and 6.3--130$\times$ on SM8850.

\subsection{Ablation Study}
\label{sec:ablations}
We ablate the two parts of our design: the selector (Section~\ref{sec:selection}) and the heterogeneous split (Sections~\ref{sec:prefill}--\ref{sec:contract}).

\subsubsection{Selector}
\label{sec:selector-ablation}
\begin{table}[!tb]
\caption{Selector ablation at matched density (Qwen3-1.7B, RULER 4k). Pooled keys improve accuracy, while pooled queries and antidiagonal scores trade additional NPU time for small gains.}
\label{tab:selector-ablation}
\vspace{0.1in}
\centering
\small
\setlength{\tabcolsep}{4pt}
\begin{tabular}{lrr}
\toprule
Selector & Score & Density (\%) \\
\midrule
\selectorablationrows
\bottomrule
\end{tabular}
\end{table}

\paragraph{Pooled keys, sampled queries.}
Table~\ref{tab:selector-ablation} changes one part of the selector at a time on Qwen3-1.7B RULER 4k, keeping the group maximum and $c^*=3$ and calibrating each variant's threshold to TileWeave's density.
Sampling keys instead of pooling them, each representative the average of two of its 16 keys, lowers the score by 0.56, most of it on UUID multi-key and multi-value retrieval (84.60 against 89.00 and 85.70 against 88.30), where two sampled keys of 16 miss the target tokens; in full-model prefill the pooled keys add only 0.5\% to prefill time at 4k and 8k.
Pooling the four query samples as well gains 0.43 points, nearly all on UUID multi-key retrieval (92.80 against 89.00), and XAttention's antidiagonal scores in the same rule gain 0.30.
Both cost more NPU time: pooling reads all 64 query rows of a block instead of four, which takes 1.5$\times$ the selection's NPU time and 3\% more prefill time at 4k and 8k, and an antidiagonal score takes 16 products instead of one, so the selector keeps sampled queries and pooled keys.

\paragraph{Selection cost.}
For Qwen3-1.7B on SM8750, we fix execution to naive 64-block attention with one list per K/V head.
Replacing XAttention's scorer with ours makes attention, selection included, \speed{1.47}/\speed{1.46} faster at 8k/16k; it is \speed{2.85}/\speed{2.83} faster than XAttention with its per-head masks, and the split adds \speed{1.52}/\speed{1.63} (Table~\ref{tab:ruler-ablation}).
End to end, the scorer makes prefill \speed{1.20}/\speed{1.27} faster and the split another \speed{1.24}/\speed{1.34}.
In a profiled prefill at 4k/8k with each group's mean, ours costs 3.4/10.0 ms of NPU time per layer over a prompt, against 9.5/27.8 ms for XAttention's scoring and cumulative selection.

\begin{table}[!tb]
\caption{Qwen3-1.7B scorer and execution ablations; gain is TileWeave's speedup over each row. The scorer and heterogeneous split make comparable contributions to the latency reduction.}
\label{tab:ruler-ablation}
\vspace{0.1in}
\centering
\footnotesize
\setlength{\tabcolsep}{2pt}
\begin{tabular}{llrrrr}
\toprule
& & \multicolumn{2}{c}{8k prompts} & \multicolumn{2}{c}{16k prompts} \\
Scorer & Execution & Time & Gain & Time & Gain \\
\midrule
\rulerablationrows
\bottomrule
\end{tabular}
\end{table}

\subsubsection{Heterogeneous Split}
\label{sec:executions}
\begin{table}[!tb]
\caption{Llama-3.2-3B executions at a fixed selector and threshold. TileWeave gives 1.11--1.18$\times$ faster prefill with competitive RULER 8k accuracy.}
\label{tab:executions}
\vspace{0.1in}
\centering
\footnotesize
\setlength{\tabcolsep}{3pt}
\begin{tabular}{lrrrrrr}
\toprule
& \multicolumn{2}{c}{RULER 8k} & \multicolumn{2}{c}{Attention (ms)} & \multicolumn{2}{c}{Prefill (s)} \\
Execution & Score & Dens. & 8k & 16k & 8k & 16k \\
\midrule
\executionrows
\bottomrule
\end{tabular}
\end{table}

\paragraph{Comparable accuracy at different densities.}
We hold the group-maximum selector and threshold fixed across the three executions on Llama-3.2-3B-Instruct and measure latency on SM8750 (Table~\ref{tab:executions}).
The masked tile retains only selected pairs (51.1\% causal-pair density), TileWeave also retains fill-in from its shared pool and head start (58.0\%), and the union retains all blocks selected within each tile (70.1\%).
On RULER 8k, TileWeave scores 0.18 points below the union and 0.34 above the masked tile; extra keys improve variable tracking but reduce common-words extraction accuracy.
An earlier group-mean control with independently calibrated thresholds compares executions at matched density (Appendix~\ref{app:executions-mean}).

\paragraph{Full-model prefill gains.}
At 8k and 16k, TileWeave accelerates full-model prefill by 1.11--1.16$\times$ over the union and 1.13--1.18$\times$ over the masked tile; the corresponding attention speedups are 1.24--1.28$\times$ and 1.29--1.34$\times$.
Both NPU-only paths compute every block selected by any of a tile's four query blocks, with the masked tile discarding fill-in.
TileWeave confines this coarse work to the shared pool and executes low-reuse blocks concurrently on the GPU, preserving the benefit through full-model prefill.

\paragraph{Reuse threshold.}
At each model's single setting, $c^*=3$ executes fewer blocks than $c^*=2$ (Appendix Table~\ref{tab:density}) and makes attention 1.09$\times$ and 1.10$\times$ faster at 8k and 16k and prefill 1.03$\times$ and 1.05$\times$ (Table~\ref{tab:ruler-ablation}), while RULER changes by $-0.51$ to $-0.12$ points with Qwen3-1.7B and $-0.25$ to $+0.15$ with Llama-3.2-3B-Instruct, and LongBench by at most 0.06.

\section{Conclusion}
TileWeave uses cross-query reuse to determine the execution granularity of sparse attention.
Highly shared key blocks run in efficient coarse NPU tiles, while low-reuse blocks run on the GPU only for the query blocks that select them.
An NPU-native selector and coordinated runtime make this partition practical.
Across two models, TileWeave maintains accuracy comparable to XAttention on RULER and LongBench.
With Qwen3-1.7B on two Snapdragon platforms, it accelerates full-model prefill by 1.35--3.23$\times$ over XAttention and up to 1.58$\times$ over dense attention.
With Llama-3.2-3B-Instruct at 8k/16k and a fixed selector and threshold, attention is 1.24--1.28$\times$ faster than the NPU-only union and 1.29--1.34$\times$ faster than masked tiles, with comparable RULER 8k accuracy.
Cross-query reuse reduces excess work while retaining efficient NPU execution.
\clearpage
\flushbottom
\bibliography{references}
\bibliographystyle{mlsys2025}
\clearpage
\raggedbottom
\section*{AI Usage Disclosure}
We used OpenAI Codex and Anthropic Claude Code as generative AI assistants during research development and manuscript preparation.
Both tools assisted with code generation and writing; Codex also assisted with literature search, experimental design, experiment execution, and result analysis.
Their use was iterative and included substantive code and prose generation as well as revision.

\paragraph{Ideation and literature.}
Codex helped refine the motivation and reasoning about selection granularity and cross-query reuse, compare alternative explanations of results, and propose follow-up controls.
For example, it assisted with designing the diagonal-preserving exception ablation reported in Table~\ref{tab:motivation-sharing}.
It also helped locate and summarize related work, compare sparse-attention methods, and check citation metadata against primary sources.

\paragraph{Code and experiments.}
Codex and Claude Code assisted with generating and revising implementation and evaluation code.
Codex's work included sparse-attention evaluation code, experiment-launch and monitoring scripts, result collection, numerical and mask validation, and debugging.
Through command-line tools, it also launched and monitored authorized GPU and phone experiments and checked saved outputs.
Validation included comparisons with reference computations, reproduction of saved baseline predictions, and checks of input, model, and configuration identities, as documented in the experimental protocols.

\paragraph{Data and analysis.}
Benchmark prompts and reference answers came from the documented RULER and LongBench datasets or their prescribed generators.
The evaluated language models generated the predictions used for accuracy scoring; latency measurements came from hardware executions.
Codex assisted with data-preparation and analysis scripts, rescoring predictions, computing aggregate scores, densities, speedups, and paired confidence intervals, and interpreting these results.
Reported numerical results were derived from saved experiment outputs, rather than generated as prose by the assistants.

\paragraph{Writing and figures.}
Both tools assisted with drafting and revising manuscript text.
Codex also helped organize and condense sections, revise captions and technical explanations, and write and modify plotting and diagram-generation scripts.
This included performance charts, method and pipeline illustrations, editable figure exports, and LaTeX layout and PDF checks.
Codex drafted this disclosure from the recorded workflow and author-provided information about tool use.
The authors directed the work and are responsible for the implementation, experimental evidence, citations, technical claims, and final manuscript.
\clearpage
\flushbottom
\appendix

\section{Per-Task Comparison}
\label{app:selector-tasks}
Tables~\ref{tab:selector-tasks} and~\ref{tab:llama-tasks} expand the RULER 4k comparison of Table~\ref{tab:ruler} for each model, Table~\ref{tab:tasks-16k} its 16k comparison for Qwen3, and Table~\ref{tab:ruler-short} adds 1k and 2k.
Each task contains the same 500 prompts in all methods, following the accuracy protocol in Section~\ref{sec:setup}.
The 1k/2k means cover 12 tasks, excluding qa\_2 because its documents do not fit these lengths.
Density counts retained causal attention pairs after the complete prefill policy, not just the selector's wanted mask; the macro density is the equal-task mean.
One setting per method leaves task densities free: at 4k they differ from TileWeave's by up to 7.2 points for XAttention and 15.1 for ShadowAttn with Qwen3, so task-level score differences cannot be attributed to ranking alone.
At 16k, with the same threshold and $c^*=3$, the group mean keeps fewer pairs than the group maximum (36.8\% against 43.8\%) and scores 68.41 against 69.99, losing most on multi-key and multi-value retrieval (43.0 against 49.2 on niah\_multikey\_2, 81.4 against 83.1 on niah\_multivalue) and on question answering (41.0 against 44.2 on qa\_1).

\begin{table*}[!t]
\caption{RULER 16k task scores and density (\%) for Qwen3-1.7B. Taking the GQA-group maximum improves retrieval over the group mean at the same threshold.}
\label{tab:tasks-16k}
\vspace{0.1in}
\centering
\small
\begin{tabular}{lrrrrrrrr}
\toprule
& \multicolumn{2}{c}{Dense} & \multicolumn{2}{c}{XAttention} & \multicolumn{2}{c}{TileWeave} & \multicolumn{2}{c}{Group mean} \\
Task & Score & Density & Score & Density & Score & Density & Score & Density \\
\midrule
\sixteenktaskrows
\bottomrule
\end{tabular}
\end{table*}

\begin{table}[!tb]
\caption{RULER 1k/2k mean scores and density (\%) for Qwen3-1.7B. TileWeave remains within 0.12 points of XAttention.}
\label{tab:ruler-short}
\vspace{0.1in}
\centering
\small
\setlength{\tabcolsep}{4pt}
\begin{tabular}{lrrrr}
\toprule
& \multicolumn{2}{c}{1k} & \multicolumn{2}{c}{2k} \\
Method & Score & Dens. & Score & Dens. \\
\midrule
\rulershortrows
\bottomrule
\end{tabular}
\end{table}

\begin{table*}[!t]
\caption{RULER 4k task scores and density (\%) for Qwen3-1.7B. TileWeave gains on multi-key retrieval but loses on question answering relative to XAttention.}
\label{tab:selector-tasks}
\vspace{0.1in}
\centering
\small
\begin{tabular}{lrrrrrrrr}
\toprule
& \multicolumn{2}{c}{Dense} & \multicolumn{2}{c}{XAttention} & \multicolumn{2}{c}{ShadowAttn} & \multicolumn{2}{c}{TileWeave} \\
Task & Score & Density & Score & Density & Score & Density & Score & Density \\
\midrule
\selectortaskrows
\bottomrule
\end{tabular}
\end{table*}

\begin{table*}[!t]
\caption{RULER 4k task scores and density (\%) for Llama-3.2-3B-Instruct. TileWeave improves variable tracking and multi-key retrieval over XAttention.}
\label{tab:llama-tasks}
\vspace{0.1in}
\centering
\small
\begin{tabular}{lrrrrrrrr}
\toprule
& \multicolumn{2}{c}{Dense} & \multicolumn{2}{c}{XAttention} & \multicolumn{2}{c}{ShadowAttn} & \multicolumn{2}{c}{TileWeave} \\
Task & Score & Density & Score & Density & Score & Density & Score & Density \\
\midrule
\llamataskrows
\bottomrule
\end{tabular}
\end{table*}

\section{Additional Selector Analysis}
\label{app:selector-analysis}

\subsection{Sharing One List across a Group's Query Heads}
\label{app:head-sharing}
The GQA kernel runs one list per K/V head, so the query heads of a group must agree on it; we compare the two rules at $c^*=2$.
Averaging their probabilities halves the score of a block that only one head needs, and at 16k this loses retrieval: on the 500 RULER 16k multi-key prompts with distractor needles (niah\_multikey\_2), the group mean scores 44.0 against XAttention's 51.8, while one list per query head scores 50.2 and the group maximum 51.0.
Swapping in XAttention's scorer (43.6) or its cumulative rule (46.0) changes little, so the gap comes from the sharing, not from the estimate.
Over all 6500 prompts the maximum adds 1.46 points at 16k (70.50 against 69.04, at 46.9\% against 39.6\% of pairs) and 0.07 at 4k (86.49 against 86.42); with Llama-3.2-3B-Instruct it adds 0.53 and 0.24 points on RULER 4k and 8k, and on LongBench it adds 0.25 points with Llama-3.2-3B-Instruct and 0.08 with Qwen3-1.7B.
It keeps a list per K/V head, so the NPU runs it with the same kernel; its extra blocks cost 8\%, 12\%, and 15\% attention time at 4k, 8k, and 16k on SM8750, and 3\%, 5\%, and 7\% prefill.

\subsection{Density at One Setting}
\label{app:density}
Each method keeps the setting calibrated at RULER 4k, so its density follows its own rule at other lengths (Table~\ref{tab:density}).
Indented rows show our $c^*=2$ controls with the group maximum and group mean.
Both block methods keep a smaller share of longer prompts, TileWeave more slowly: from 1k to 16k our share of causal pairs falls from 82.1\% to 43.8\% and XAttention's from 86.3\% to 36.2\%.
Our threshold was calibrated with the group mean at $c^*=2$, which keeps 53.2\% at 4k; the group maximum keeps 60.4\% at the same threshold, and 57.3\% with $c^*=3$.
On the phone XAttention executes fewer blocks than TileWeave at 1k--8k, yet its attention is 2.2--3.0$\times$ slower than dense attention (Section~\ref{sec:performance}), because it runs them as per-head 64-query tiles.

\begin{table}[H]
\caption{Density (\%) of retained causal pairs and executed phone blocks for Qwen3-1.7B (LB: LongBench). Both block methods become sparser on longer prompts, TileWeave more slowly.}
\label{tab:density}
\vspace{0.1in}
\centering
\small
\setlength{\tabcolsep}{2.5pt}
\begin{tabular}{llrrrrrr}
\toprule
& Method & 1k & 2k & 4k & 8k & 16k & LB \\
\midrule
\densityrows
\bottomrule
\end{tabular}
\end{table}

\section{Executions at One Density}
\label{app:executions-mean}
Before the group maximum, we compared the three executions with each GQA group's mean selections and gave each execution its own threshold for the same 53\% density (Table~\ref{tab:executions-mean}; Llama-3.2-3B-Instruct, attention of the sparse chunks on SM8750 with the optimized split operator of that study).
At that density the union scores 85.12 on RULER 4k, a point below TileWeave with $c^*=3$ (86.10) and the masked tile (86.15), which preserve each query block's own selection; common-words extraction carries most of the loss, 23.8 for the union against 32.3 with $c^*=3$ and 33.1 for dense attention.
With $c^*=2$, every block that two of the four query blocks request stays in the shared tile, so it carries much of the same fill-in: 85.64, and 24.4 on common words.
TileWeave is the fastest of the three: the union takes 1.14$\times$ (4k) and 1.12$\times$ (8k) its attention time, and the masked tile, which executes every block its four query blocks select (69\% and 65\% of causal blocks at 4k and 8k), 1.45$\times$ and 1.54$\times$.

\begin{table}[H]
\caption{Llama-3.2-3B at matched density: TileWeave matches masked-tile accuracy and reduces attention latency. Parentheses show latency relative to TileWeave.}
\label{tab:executions-mean}
\vspace{0.1in}
\centering
\small
\setlength{\tabcolsep}{3pt}
\begin{tabular}{lrrrr}
\toprule
& \multicolumn{2}{c}{RULER 4k} & \multicolumn{2}{c}{Attention (ms)} \\
Execution & Score & Dens. & 4k & 8k \\
\midrule
TileWeave, $c^*=3$ & 86.10 & 53.1 & 374 & 1277 \\
Masked tile & 86.15 & 53.1 & 542 (1.45) & 1973 (1.54) \\
Union & 85.12 & 53.2 & 426 (1.14) & 1426 (1.12) \\
\bottomrule
\end{tabular}
\end{table}

\section{Implementation and Measurement Details}
\subsection{ShadowAttn Execution}
\label{app:shadowattn-execution}
We reimplement ShadowAttn's per-query-row selection with one retention ratio per model, a policy also evaluated in its published ablation \citep{yin2026shadownpu}.
The accuracy reference quantizes Q/K to INT8 and ranks their integer products.
For phone timing, the NPU instead computes F16 HMX products and emits F32 scores; the CPU or GPU maps each row's scores to 8 bits before selecting its highest-scoring keys, with ties broken by key index.
These estimation paths share the selection rule but need not select identical keys.
Estimation precedes selection and attention in the model graph; we do not reproduce the published pipeline across heads.

The CPU version gathers only selected K/V rows and computes attention with F16 products and F32 accumulation.
The GPU version uses OpenCL selection followed by masked FlashAttention with F16 K/V and F32 accumulation: it computes causal tiles and masks unselected pairs, retaining dense tile computation.
On SM8850, a diagnostic using the first variable-tracking prompt per length found that only 0.27\%, 0.45\%, and 1.05\% of causal $64\times32$ query-key tiles contain no selected pairs at 4k, 8k, and 16k.
A prototype GPU kernel that gathers individual selected tokens had 9.3$\times$ and 17.0$\times$ higher attention-kernel latency than the masked kernel at 4k and 8k in this diagnostic.
We therefore report the faster masked implementation; the results characterize our implementations under the stated model, precision, and resource settings.

\subsection{Attention Timing}
\label{app:attention-timing}
We directly time each chunk's attention graph across all layers using captured queries and resident K/V, including estimation, selection, attention, and device coordination and merging.
Q/K/V projections, cache writes, and the other model operators are outside this timer.
Each prompt's attention latency is the sum over its chunks; ShadowAttn's inputs are captured during dense prefill.
Full-model prefill is timed separately with the complete model graph, so Figures~\ref{fig:attention-latency} and~\ref{fig:prefill-latency} are separate measurements rather than an additive latency breakdown.
For example, on SM8750, ShadowAttn's CPU attention at 4k takes 18.42\,s when measured separately, while its full-model prefill takes 15.36\,s.
Same-device checks find that CPU attention chunks with 2--4k cached keys take 1.3--1.5$\times$ longer when measured separately than within prefill; later chunks agree within 1\%.
We retain both measurements and use the full-model prefill timings for end-to-end speedup claims.

\subsection{Runtime Timing and Publication Diagnostics}
\label{app:runtime-diagnostics}
These diagnostics motivated the runtime in Section~\ref{sec:contract}.
They come from historical execution studies and isolated harnesses, rather than the current full-model performance cohort; their costs are not additive.

\paragraph{Pipeline calibration.}
Figure~\ref{fig:pipeline} covers 1024 queries and eight K/V heads, with endpoints set to two-round means from the historical 4k Qwen3-1.7B/SM8750 study using the earlier threshold scorer.
Selection finishes before the illustrated interval; other model operators are excluded.
Tile intervals are approximate, and store/merge marks encode no duration.
GPU completion spans 0.37--1.47\,ms from a nearby trace, with intermediate groups interpolated.
The figure illustrates overlap and relative latency rather than a single event trace.

\paragraph{Startup and progressive completion.}
The GPU's first partial is ready 0.35--0.48\,ms after the NPU publishes the operator's inputs.
A conventional launch over all exceptions finishes the first of its 32 stages only after 2.5\,ms of a 3.9\,ms execution; one launch per stage instead adds about 65\us at each boundary.
Persistent workgroups publish successive stages about 27--30\us apart in the recorded trace.
These observations motivate the eight-tile NPU head start and progressive publication in consumption order.

\paragraph{Merge placement and memory traffic.}
A separate vector pass over the NPU output adds 890--920\us per operator, exceeding the historical split's 0.67--0.80\,ms saving over naive 64-block attention at 4k and 8k.
In the isolated fold harness, changing the HMX normalization adds 290--350\us, while merging in the existing store with tile-ahead prefetch adds 113--186\us; omitting that prefetch raises the overhead to about 980\us.
The GPU writes F16 partials to halve their transfer size relative to F32, reads K/V as 16-byte texels, and reads Q through an image.
In a separate 8k profile, GPU exceptions take about 2.2\,ms per operator and expose 218\us of NPU wait; NPU matrix multiplication is 7\% slower with concurrent GPU traffic.

\paragraph{Payload visibility.}
With only an atomic ready word, pre-enqueued GPU work read the previous layer's queries from a reused buffer in 10--20\% of checked rows; fences, atomic loads, and added delays did not prevent it.
The host relay's buffer-referencing submission makes the GPU observe current Q and membership data in the tested driver path.
NPU readers invalidate published rows unless a whole-L2 flush already did so, and prefetch exact row ranges so that a future stage is not cached before the GPU writes it.
These mechanisms describe the tested shared-buffer implementation, not a portable guarantee for arbitrary OpenCL aliases.


\end{document}